%% file: main.tex
\documentclass[runningheads]{llncs}
\usepackage{graphicx}
\usepackage{amsmath,amssymb} 
\usepackage{color}

\usepackage[caption=false]{subfig}
\usepackage{stackengine}

\newsavebox\CBox
\def\textBF#1{\sbox\CBox{#1}\resizebox{\wd\CBox}{\ht\CBox}{\textbf{#1}}}
\newcommand{\bestResult}[1]{\textBF{#1}}

\newcommand{\zht}[1]{#1}
\newcommand{\zhtdel}[1]{}

\newcommand{\smallSection}[1]{\noindent{\textbf{#1}}}

\begin{document}
\title{CrossNet: An End-to-end Reference-based Super Resolution Network using Cross-scale Warping} 

\titlerunning{Learning the reference-based super resolution}

\author{Haitian Zheng\inst{1}
Mengqi Ji\inst{1,2}
Haoqian Wang\inst{1}
Yebin Liu\inst{3}
Lu Fang\inst{1} \thanks{The correspondence author is Lu FANG (\email{fanglu@sz.tsinghua.edu.cn}). This work is supported in part by Natural Science Foundation of China (NSFC) under contract No. 61722209, 61331015 and 61522111, in part by the National key foundation for exploring scientific instrument of China No.2013YQ140517.}
}
%
\authorrunning{H. Zheng, M. Ji, H. Wang, Y. Liu, L. Fang}
%

\institute{Tsinghua University, Tsinghua-Berkeley Shenzhen Institute  \and
Hong Kong University of Science and Technology \and
Tsinghua University, Dept. of Automation
}
\maketitle              
\begin{abstract}
The Reference-based Super-resolution (RefSR) super-resolves a low-resolution (LR) image given an external high-resolution (HR) reference image, where the reference image and LR image share similar viewpoint but with significant resolution gap ($8\times$). Existing RefSR methods work in a cascaded way such as patch matching followed by synthesis pipeline with two independently defined objective functions, leading to the inter-patch misalignment, grid effect and inefficient optimization. \zht{To resolve these issues, we present CrossNet, an end-to-end and fully-convolutional deep neural network using cross-scale warping. Our network contains image encoders, cross-scale warping layers, and fusion decoder: the encoder serves to extract multi-scale features from both the LR and the reference images; the cross-scale warping layers spatially aligns the reference feature map with the LR feature map; the decoder finally aggregates feature maps from both domains to synthesize the HR output. Using cross-scale warping, our network is able to perform spatial alignment at pixel-level in an end-to-end fashion, which improves the existing schemes \cite{PatchMatch,SS-Net} both in precision (around 2dB-4dB) and efficiency (more than 100 times faster).} 


\keywords{Reference-based Super resolution \and Light field imaging \and Image synthesis \and Encoder-decoder \and Optical flow}
\end{abstract}

\input{section/intro}
\input{section/related}
\input{section/model_v3}

\input{section/experiment}

\input{section/conclude}

\bibliographystyle{splncs}
\bibliography{egbib}
\end{document}

%% file: section/intro.tex
\section{Introduction}
Reference-based super-resolution (RefSR) methods \cite{SS-Net} utilizes an extra high resolution (HR) reference image to help super-resolve the low resolution (LR) image that shares similar viewpoint. Benefit from the high resolution details in reference image, RefSR usually leads to competitive performance compared to single-image SR (SISR). While RefSR has been successfully applied in light-field reconstruction \cite{PatchMatch,SR_lr_TVCG16,SS-Net} and giga-pixel video synthesis \cite{giga}, it remains a challenging and unsolved problem, due to the parallax and the huge resolution gap (8x) exist between HR reference image and LR image. Essentially, how to transfer the high-frequency details from the reference image to the LR image is the key to the success of RefSR. This leads to the two critical issues in RefSR, i.e., image correspondence between the two input images and high resolution synthesis of the LR image. 

\vspace{-0.3cm}
\begin{figure}
	\centering
	\includegraphics[width=1.0\textwidth]{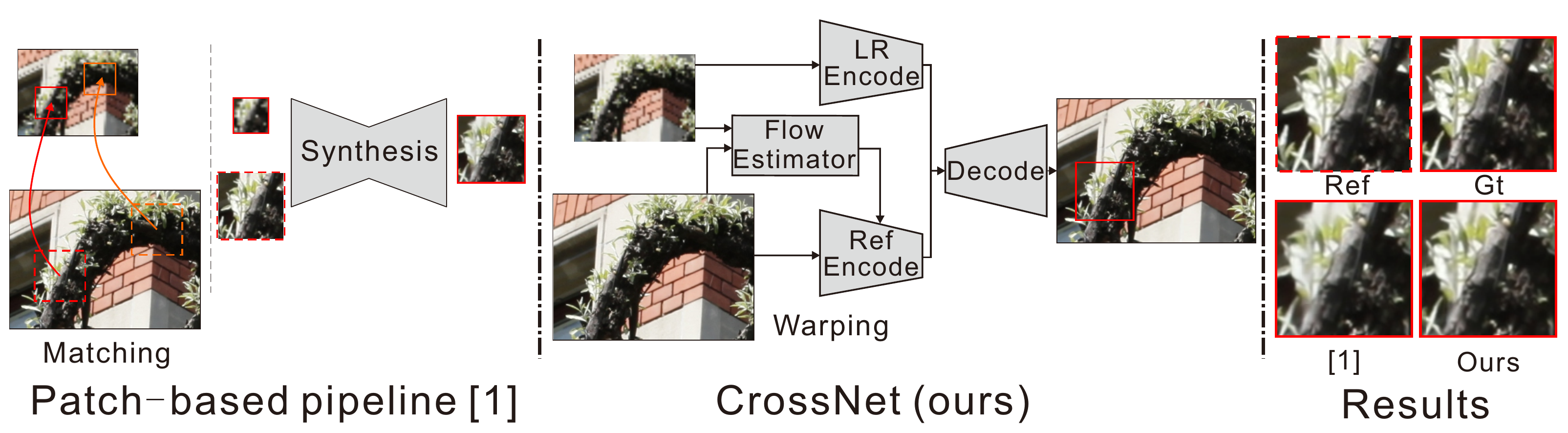}
	\caption{ Left: the `patch maching + synthesis' pipeline of \cite{SS-Net}, middle: the proposed end-to-end CrossNet, right: results comparisons.
	}
	\label{fig:teaser}
\end{figure}
\vspace{-0.5cm}

In the initial work of \cite{PatchMatch}, to develop image correspondences between the two inputs, the gradient features on the down-sampled patches in the HR reference are used for patch-based matching, while patch averaging is designed for the image synthesis. However, the oversimplified and down-sampled correspondence estimation of \cite{PatchMatch} does not take advantage of the high frequency information for matching, while the synthesizing step does not utilize high resolution image prior for better fusion. To address the above two limitations, a recent work \cite{SS-Net} replaces the gradient feature of \cite{PatchMatch} with the convolutional neural network (CNN) learned features to improve the matching accuracy, and then proposes an additional CNN which utilizes the state-of-the-art single image super-resolution (SISR) algorithm \cite{srgan} for patch synthesis. However, the `patch matching + patch synthesis' scheme of \cite{PatchMatch,SS-Net} are fundamentally limited. Firstly, the adopted sliding averaging blurs the output image and causes grid artifacts. Moreover, patch-based synthesis is inherently incapable in handling the non-rigid image deformation caused by viewpoint changes. To impose the non-rigid deformation to patch-based algorithms, \cite{SR_lr_TVCG16} enriches the reference images by iteratively applying non-uniform warping before the patch synthesis. However, directly warping between the low and high resolution images is inaccurate. In addition, such iterative combination of patch matching and warping introduces heavy computational burden, e.g. around 30min for synthesizing an image.

In this paper, we propose CrossNet, an end-to-end convolutional neural network based on the idea of `warping + synthesis' for reference-based image super-resolution. We discard the idea of `patch matching' and replace it with `warping', which enables the design of `Encoder-Warping-Decoder' structure, as shown in Fig. \ref{fig:teaser}. Such structure contains two encoders to extract multi-scale features from LR and reference image respectively. We take advantage of the warping module originated from spatial transformer network (STN) \cite{STN}, and integrate it to our HR reference image encoder. Compared with the patch matching based methods, warping naturally supports non-rigid deformation to overcome the parallax challenge in RefSR. More over, we extract multi-scale features in the encoder, and then perform multi-scale spatial alignment using warping, as shown in Fig.~\ref{fig:teaser}. The introduced multi-scale features capture the complementary scale information from two images, which help to alleviate the huge resolution gap challenge in RefSR. Finally, the decoder aggregates features to synthesize the HR output. Overall, our model is fully end-to-end trainable and does not require pretraining the flow estimator.

Extensive experiments have shown the superior performance of CrossNet (around 2dB-4dB gain) compared to state-of-the-art SISR and RefSR methods, under different datasets with large/small viewpoint disparities and different scales. Our trained model that generalized to external dataset including Stanford light field maintains the ability to retain high frequency details. More importantly, CrossNet is efficient in terms that it generates a $320\times512$ image within one second, while \cite{PatchMatch}, \cite{SS-Net} and \cite{SR_lr_TVCG16} take 86.3s, 105.0s and around 30 minutes to perform the same task, respectively. 

%
%
%

%% file: section/related.tex
\vspace{-8pt}
\section{Related work}
\vspace{-5pt}
\subsection{Single-image Super-resolution}
The single-image super-resolution (SISR) problem aims to super-resolve an LR image without additional references. Despite that, the SISR problems are closely related to the Reference-based Super-resolution (RefSR) problem. In the early days, approaches based on adaptive sampling \cite{interpolation_SR1,interpolation_SR2} has been applied to SISR. However, such approaches did not utilize the statistics of nature images. In contrast, model-based approaches try to design image prior which helps to super-resolve the image-specific patterns. Such works usually utilize edge prior \cite{edge_prior_SR}, total variation model \cite{TV_SR}, hyper-Laplacian prior \cite{hyper-Laplacian-SR}, sparsity priors \cite{sparse_SR1,sparse_SR2,sparse_SR3,sparse_SR4}, or exemplar patches \cite{innter_exemplar_SR1,exemplar_SR5}.

More recently, the SISR problem was casted into a supervised regression problem, which try to learn a mapping function from LR patches to HR patches. Those works relies on varieties of learning techniques including nearest-neighbor search \cite{exemplar_SR2_LLE,NLM}, decision tree \cite{SR_tree}, random forests \cite{exemplar_SR6_forests,SR_tree2}, simple function \cite{anchored_simplefun,exemplar_SR1_simplefun}, Gaussian process regression \cite{SR_GPR}, and deep neural networks.

With the increasing model capacity of the deep neural networks, the SISR performance has been rapidly improved. Since the appearance of the first deep learning-based SR method \cite{SRCNN}, a large number of works have been proposed to further improve the SISR performance. For example, Dong et al.~\cite{FSRCNN} and Shi et al.~\cite{subpixel} accelerate the efficiency of SISR by computing features on low-resolution domains. Kim et al.~\cite{VDSR} proposed a 20-layers deep network for predicting the bicubic upsampling residue. Ledig et al.~\cite{srgan} proposed a deep residue network with adversarial training for SISR. Lai et al.~\cite{deep_laplacian_SR} reconstructed the sub-band residuals using a multi-stage Laplacian network. Lim et al.~\cite{MDSR} improved \cite{srgan} by introducing a multi-scale feature extraction residue block for better performance. Because of the impressive performance of the MDSR network from \cite{MDSR}, we employ MDSR as a sub-module for LR images feature extraction and RefSR synthesis.
\vspace{-8pt}
\subsection{Reference-based Super-resolution}
Recent works such as \cite{SS-Net,PatchMatch,SR_lr_TVCG16,SR_lf_ECCV2012,SR_lr_CVPRW12,SR_lf_VCIP15} uses additional reference images from different viewpoints to help super-resolving the LR input, which forms a new kind of SR method called RefSR. Specifically, Boominathan et al.~\cite{PatchMatch} used an DSLR captured high-definition image as reference, and applies a patch-based synthesizing algorithm using non-local mean \cite{NLM} for super-resolving the low-definition light-field images. Wu et al.~\cite{SR_lf_VCIP15} improved such algorithm by employing patch registration before the nearest neighbor searching, then applies dictionary learning for reconstruction. Wang et al.~\cite{SR_lr_TVCG16} iterate the patch synthesizing step of \cite{PatchMatch} for enriching the exemplar database. Zheng et al.~\cite{ICCVW2017} decompose images into subbands by frequencies and apply patch matching for high-frequency subband reconstruction. Recently, Zheng et al.~\cite{SS-Net} proposed a deep learning-based approach for the cross-resolution patch matching and synthesizing, which significantly boosts the accuracy of RefSR. However, the patch-based synthesizing algorithms are inherently incapable in handling the non-rigid image deformation that is often caused by the irregular foreground shapes. \zht{Under such cases, patch-based synthesize causes issues such as blocky artifact and blurring effect.}
Despite that sliding windows \cite{PatchMatch,SS-Net} or iterative refinement \cite{SR_lr_TVCG16} mitigate such difficulties to some extends, these strategies usually introduce heavy computational cost.
On the contrary, our fully convolutional network makes it possible to achieve more than 100 times speedup compared to existing RefSR approaches, allowing the model to be applicable for real-time applications.
\par


\zht{
\vspace{-8pt}
\subsection{Image/video Synthesis using Warping}
Our task is also related to image/video synthesis tasks that use additional images from other viewpoints or frames. 
Such tasks include view synthesis \cite{Learning-based-view-synthesis,deep_view_morphing}, video denoising \cite{task_flow}, super-resolution \cite{task_flow,video-super-resolution1,video-super-resolution2}, interpolation or extrapolation \cite{deep_voxel_flow,LFVideo}. 
To solve this type of problems, deep neural networks based of the design of ``warping and synthesis'' has been recently proposed. Specifically, the additional images are backward warped to the target image using the estimated flow. Afterward, the warped image is used for image/frame synthesis using an additional synthesis module. We follow such ``warping and synthesis'' pipeline. However, our approach is different from existing works in the following ways: 1) in stead of the common practice where warping was performed on image-domain at pixel-scale \cite{deep_voxel_flow,LFVideo,Learning-based-view-synthesis,deep_view_morphing}, our approach performs multi-scale warping on feature domain, which accelerates the model convergence by allowing flow to be globally updated at higher scales. 2) after the warping operations, a novel fusion scheme is proposed for image synthesis. Our fusion scheme is different from the existing synthesizing practices that include image-domain early fusion (concatenation) \cite{deep_voxel_flow,deep_view_morphing} and linearly combining images \cite{LFVideo,Learning-based-view-synthesis}.
} 

%% file: section/model_v3.tex
\vspace{-8pt}
\section{Approach}
\begin{figure}
	\centering
	\includegraphics[width=0.7\textwidth]{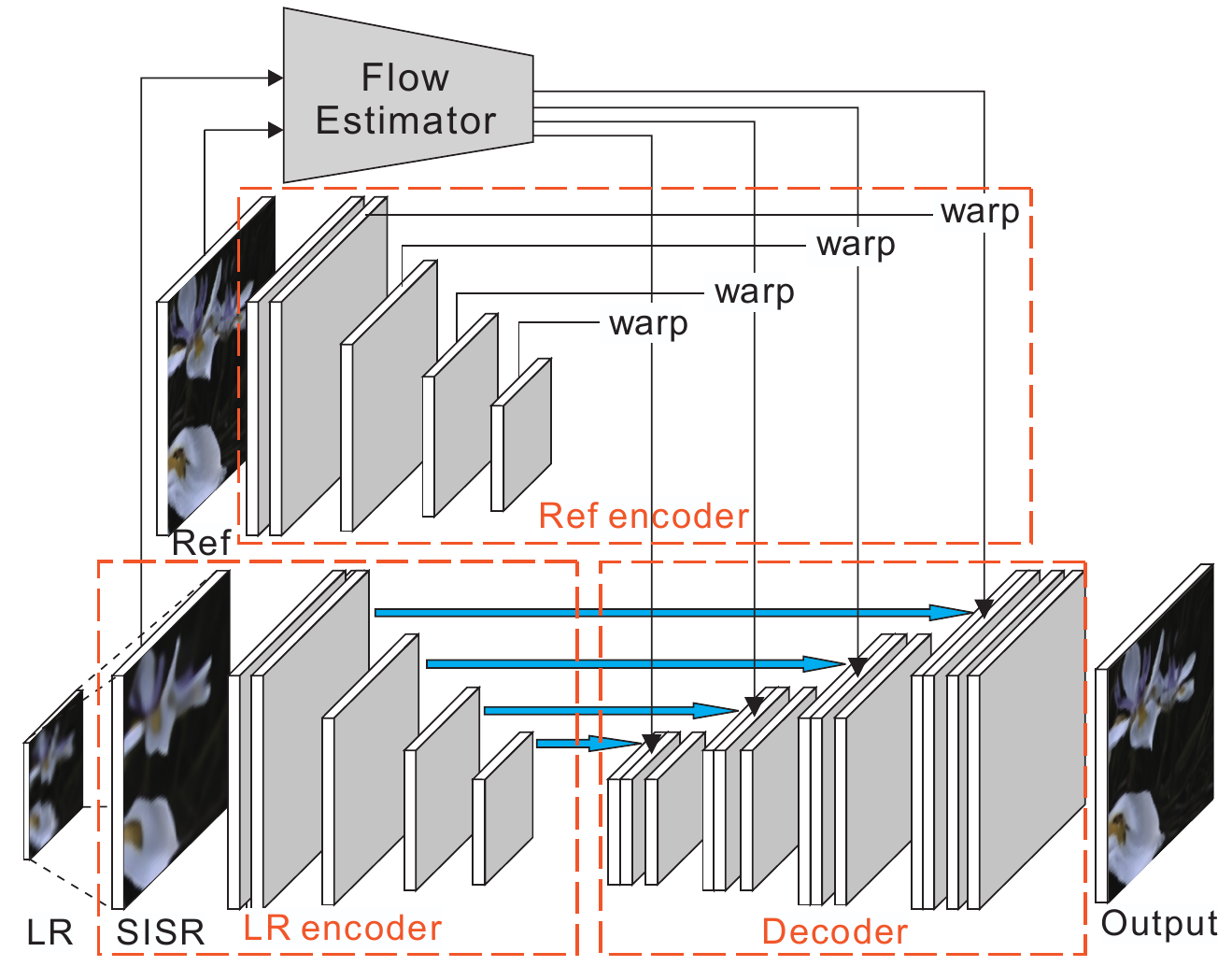}
	\caption{Network structure of our proposed CrossNet.}
	\label{fig:model}
\end{figure}



Our reference-based super resolution scheme, named CrossNet, is based on a fully convolutional cross-scale alignment module that spatially aligns the reference image information to the LR image domain. Along with the cross-scale alignment module, an encoder-decoder structure is proposed to directly synthesize the RefSR output in an end-to-end, and fully convolutional fashion. The entire network is plotted in Fig. \ref{fig:model}. In Section 3.1, we introduce the designs and properties of the fully convolutional cross scale alignment module. In Section 3.2, the end-to-end network structure is described, followed by the image synthesis loss function depicted in Section 3.3.

\vspace{-8pt}
\subsection{Fully Convolutional Cross-scale Alignment Module}

Since the reference image is captured at different view points from LR image, it is necessary to perform spatial alignment. In \cite{PatchMatch,SS-Net,SR_lr_TVCG16}, such correspondence is estimated by matching every LR patches with its surrounding reference patches. However, such sparsely-sampled and non-rigidly upsampled correspondence can easily fail around the region with varying depth or disparity.

\smallSection{Cross-scale warping.} We propose cross-scale warping to perform non-rigid image transformation. Comparing to patch matching, we do not assume the depth plane to be locally constant. Our proposed cross-scale warping operation considers a pixel-wise shift vector $V$:
\begin{eqnarray}
I_o = warp(y_{Ref}, V),
\end{eqnarray}
which assigns a specific shift vector for each pixel location, so that it avoids the blocky and blurry artifacts.
\par

\smallSection{Cross-scale flow estimator.} 
As shown on the top of Fig. \ref{fig:model}, given an upsampled LR image
and its corresponding reference image, we adopt the widely used FlowNetS \cite{FlowNet} as our flow estimator to generate the cross-scale correspondence at multiple scale. To further improve the FlowNetS,
we replace the final $\times 4$ bilinear upsampling layer of FlownetS with two $\times2$ upsampling module, whereas each $\times2$ upsampling module contains a skip connection structure following a deconvolution layer. Such additional upsampling procedure allow the modified model to predict the flow-field with much finer definition. The modified flow estimator works to generate the multi-scale flow-fields as follows:
\begin{eqnarray}
\{V^{(3)}, V^{(2)}, V^{(1)}, V^{(0)}\} = {Flow}(I_{LR\uparrow}, I_{REF}),
\label{equ:multiscale}
\end{eqnarray}
where the $I_{REF}$ denotes the reference image, and $I_{LR\uparrow}$ denotes an representative Single-Image SR (SISR) approach \cite{MDSR} upsampled the LR image ($I_{LR}$):

\begin{equation}
\begin{array}{rl}
I_{LR\uparrow}&= SISR(I_{LR}). \\
\end{array}
\label{equ:SISR}
\end{equation}

More discussions on the choice of flow estimator are presented in discussion in Section \ref{Sec:ExpDis}.

\vspace{-8pt}
\subsection{End-to-end Network Structure}
The patch matching calculates pixel-wise flow using a sliding window scheme. Such matching is computationally expensive, compared with the proposed fully convolutional network for cross-scale flow field estimation.

Resorting the cross-scale warping as a key component for spatial alignment, we propose an end-to-end network for RefSR synthesis. Our network, contains a  \textbf{LR image encoder} which extracts multi-scale feature maps from the LR image $I_L$, a \textbf{reference image encoder} which extracts and aligns the reference image feature maps at multiple scales
, and a \textbf{decoder} which perform multi-scale feature fusion and synthesis using the U-Net\cite{U-Net} structure. Fig. \ref{fig:model} summarizes the structure of our proposed CrossNet. The major modules, i.e., encoder, estimator and decoder, are elaborated as follows.

\smallSection{LR image encoder.} Given the LR image $I_L$, we design a LR image encoder to extract reference feature maps at 4 scales. Specifically, we utilize SISR approach in Equation \ref{equ:SISR} to upsample the LR image. After that, we convolve the upsampled images with 64 filters (of size $5 \times 5$) with stride 1 to extract feature map at scale $0$. We repeatedly convolve the feature map at the scale $i-1$ (for $0 < i \le 3$) with 64 filters (of size $5 \times 5$) with stride 2 to extract feature map at scale $i$. Such operations can be represented as
\begin{equation}
\begin{array}{rl}
F^{(0)}_{LR} &= \sigma(\boldsymbol{W}^{(0)}_{LR} \ast I_{LR\uparrow} + \boldsymbol{b}^{(0)}_{LR}), \vspace{0.2cm}\\
F^{(i)}_{LR} &= \sigma(\boldsymbol{W}^{(i)}_{LR} \ast F^{(i-1)}_{LR} + \boldsymbol{b}^{(i)}_{LR}) {\Downarrow_2}, ~ i = {1,2,3},
\end{array}
\label{equ:LR_encoder}
\end{equation}
where $F^{(i)}_{LR}$ is the LR feature map at scale $i$, $\sigma$ stands for the activation function of rectified linear unit (ReLU) \cite{RELU}, $\ast$ denotes convolution, and $\Downarrow_2$ denotes 2D sampling with stride 2.  

Note that resorting independent SISR approaches (such as \cite{MDSR} ) to encode LR image owns two advantages. First, the SISR approaches that are validated on large-scale external datasets help the LR image encoder to generalize better on unseen scenes. Second, new state-of-the-art SISR approaches can be easily integrated into our system to improve the performance without changing our network structures.

\smallSection{Reference image encoder.} Given the raw reference image $I_R$, a 4 scale feature extraction network with the exact structure from Equation \ref{equ:LR_encoder} are used to sequentially extract reference image features $\{F^{(0)}_{REF},F^{(1)}_{REF},F^{(2)}_{REF},F^{(3)}_{REF}\}$ from multiple scales.
We allow the reference feature extractor and the LR feature extractor to learn different weights, which helps the two sets of features to complement each other. 

After that, we perform backward warping operation on the reference image features $F^{(i)}_R$ using the cross-scale flow $V^{(i)}$ in equation \ref{equ:multiscale}, to generate the spatially aligned feature $\hat{F}^{(i)}_{R}$.
\begin{eqnarray}
\hat{F}^{(i)}_{REF} = warp(F^{(i)}_{REF}, V^{(i)}), ~ i = {0,1,2,3}.
\end{eqnarray}
More discussions on the multi-scale warping are presented in Section \ref{Sec:ExpDis}.

\smallSection{Decoder.}
After extracting the LR image feature and the warped reference image feature at different scales, a U-Net like decoder is proposed to perform fusion and SR synthesis. Specifically, the warped features and the LR image features at scale $i$ (for $0 \le i \le 3$), as well as the decoder feature from scale $i-1$ (if any) are concatenated following a deconvolution layer with 64 filters (of size $4 \times 4$) and stride 2 to generate decoder features at scale $i$,
\begin{equation}
\begin{array}{rl}
F^{(3)}_{D} &= \sigma(\boldsymbol{W}^{(3)}_{D} \star (F^{(3)}_{LR}, \hat{F}^{(3)}_{REF}) + \boldsymbol{b}^{(3)}_{D}), \\\\
F^{(i)}_{D} &= \sigma(\boldsymbol{W}^{(i)}_{D} \star (F^{(i+1)}_{LR}, \hat{F}^{(i+1)}_{REF}, F^{(i+1)}_{D}) + \boldsymbol{b}^{(i)}_D), ~ i = {2,1,0},
\end{array}
\end{equation}
where $\star$ denotes the deconvolution operation.

After generating the decoder feature at scale $0$, three additional convolution layers with filter sizes $5 \times 5$ and filter number $\{64, 64, 3\}$ are added to perform post-fusion and to generate the SR output,
\begin{equation}
\begin{array}{lll}
F^{}_{1} & = &\sigma(\boldsymbol{W}^{}_{1} \ast F^{(0)}_{D} + \boldsymbol{b}^{}_{1}), \\
F^{}_{2} &=& \sigma(\boldsymbol{W}^{}_{2} \ast F^{}_{1} + \boldsymbol{b}^{}_{2}),	\\
I_p &=& \sigma(\boldsymbol{W}^{}_{p} \ast F^{}_{2} + \boldsymbol{b}^{}_{p}).
\end{array}
\end{equation}

\subsection{Loss Function}
Our network can be directly trained to synthesize the SR output. Given the network prediction $I_{p}$, and the ground truth high-resolution image $I_{HR}$, the loss function can be written as
\begin{eqnarray}
\mathcal{L} = \frac{1}{N} \sum_{i=1}^{N} \sum_{s} \rho (I_{HR}^{(i)}(s) - I_{p}^{(i)}(s) ),
\end{eqnarray}
where $\rho(x)=\sqrt{x^2+0.001^2}$ is the Charbonnier penalty function \cite{Charbonnier}, N is the number of samples, $i$ and $s$ iterate over training samples and spatial locations, respectively.

%% file: section/experiment.tex
\vspace{-8pt}
\section{Experiment}

\vspace{-8pt}
\subsection{Dataset}
The representative Flower dataset \cite{Flower} and Light Field Video (LFVideo) dataset \cite{LFVideo} are used here. The Flower dataset \cite{Flower} contains $3343$ flowers and plants light-field images captured by Lytro ILLUM camera, whereas each light field image has $376 \times 541$ spatial samples, and $14 \times 14$ angular samples. Following \cite{Flower}, we extract the central $8 \times 8$ grid of angular sample to avoid invalid images, and randomly divide the dataset into $3243$ images for training and $100$ images for testing. The LFVideo dataset \cite{LFVideo} contains real-scene light-field image captured by Lytro ILLUM camera. Similar to the Flower dataset, each light field image has $376 \times 541$ spatial samples and $8 \times 8$ angular samples. There are in total 1080 light-field samples for training and 270 light-field samples for testing.

For model training using these two dataset, the LR and reference images are randomly selected from the $8 \times 8$ angular grid. While for testing, the LR images at angular position $(i, i), 0 < i \le 7$ and reference images at position $(0, 0)$ are selected for evaluating RefSR algorithms. As our model requires the input size being a factor of 32, the images from the two dataset are cropped to $320 \times 512$ for training and validation. 

To validate the generalization ability of CrossNet, we also test it on the images from Stanford Light Field dataset \cite{Stanford-light-field} and Scene Light Field dataset \cite{Scene_LF}, where we apply our trained model using sliding windows approach, with windows size being $512\times512$ and stride being $256$ to output the SR result of the entire image. More details are presented in the generalization analysis in \ref{subsection:CrossNet}.

\vspace{-8pt}
\subsection{Evaluation}
\label{subsection:CrossNet}
We train the CrossNet for 200K iterations on the Flower and LFVideo datasets for $\times 4$ and $\times 8$ SR respectively. The learning rates are initially set to 1e-4 and 7e-5 for the two dataset respectively, and decay to 1e-5 and 7e-6 after 150k iterations. As optimizer, the Adam \cite{ADAM} is used with $\beta_1 = 0.9$, and $\beta_1 = 0.999$. In comparison to CrossNet, we also test the latest RefSR algorithms SS-Net \cite{SS-Net} and PatchMatch \cite{PatchMatch}, and the representative SISR approaches including SRCNN \cite{SRCNN}, VDSR \cite{VDSR} and MDSR \cite{MDSR}. 

We evaluate the results using three image quality metrics: PSNR, SSIM \cite{SSIM}, and IFC \cite{IFC}.
Table \ref{Table:main_experiment} shows quantitative comparisons for $\times 4$ and $\times 8$ RefSR under the two parallax settings, where the reference images are sampled at position $(0,0)$ while LR images are sampled at position $(1,1)$ and $(7,7)$. Examining Table \ref{Table:main_experiment}, the proposed CrossNet outperforms the previous approaches considerably under various settings including small/large parallax, different upsampling scales and different datasets, achieving 2dB-4dB gain in general. 

\begin{table*}[htbp]
	\resizebox{\textwidth}{!}{%
	\centering
	\begin{tabular}{ |l|c|c|c|c|c| }
		\hline
		Algorithm 						& Scale 	& Flower (1,1)	& Flower (7,7) & LFVideo (1,1)& LFVideo (7,7) \\
		 								& 			& PSNR/SSIM/IFC & PSNR/SSIM/IFC & PSNR/SSIM/IFC & PSNR/SSIM/IFC\\
		\hline
		SRCNN \cite{SRCNN}				& $\times 4$&32.76 / 0.89 / 2.46 &32.96 / 0.90 / 2.49 &32.98 / 0.86 / 2.07& 33.27 / 0.86 / 2.08\\
		VDSR \cite{VDSR}				& $\times 4$&33.34 / 0.90 / 2.73 &33.58 / 0.91 / 2.76 &33.58 / 0.87 / 2.29& 33.87 / 0.88 / 2.30\\	
		MDSR \cite{MDSR}				& $\times 4$&34.40 / 0.92 / 3.04 &34.65 / 0.92 / 3.07 &34.62 / 0.89 / 2.62& 34.91 / 0.90 / 2.63\\					
		PatchMatch \cite{PatchMatch}	& $\times 4$&38.03 / 0.97 / 5.11 &35.23 / 0.94 / 3.85  &38.22 / 0.95 / 4.60& 37.08 / 0.94 / 3.99\\
		\bestResult{CrossNet (ours)}			& $\times 4$&\bestResult{42.09 / 0.98 / 6.70} &\bestResult{38.49 / 0.97 / 5.02} &\bestResult{42.21 / 0.98 / 5.96}& \bestResult{39.03 / 0.96 / 4.61}\\
		\hline
		SRCNN \cite{SRCNN}				& $\times 8$&28.17 / 0.77 / 0.98&28.25 / 0.77 / 1.00&29.43 / 0.75 / 0.82&29.63 / 0.76 / 0.82 \\
		VDSR \cite{VDSR}				& $\times 8$&28.58 / 0.78 / 1.04&28.68 / 0.78 / 1.06&29.83 / 0.77 / 0.89&30.04 / 0.77 / 0.89\\	
		MDSR \cite{MDSR}				& $\times 8$&29.15 / 0.79 / 1.17&29.26 / 0.80 / 1.19&30.43 / 0.78 / 1.04&30.65 / 0.79 / 1.05 \\					
		PatchMatch \cite{PatchMatch}	& $\times 8$&35.26 / 0.95 / 4.00&30.41 / 0.85 / 2.07&36.72 / 0.94 / 3.81&34.48 / 0.91 / 2.84 \\
		SS-Net \cite{SS-Net}& $\times 8$&37.46 / 0.97 / 4.72&32.42 / 0.91 / 2.95&37.93 / 0.95 / 4.06&35.81 / 0.93 / 3.30 \\
		\bestResult{CrossNet (ours)}		& $\times 8$&\bestResult{40.31 / 0.98 / 5.74}&\bestResult{34.37 / 0.93 / 3.45 }&\bestResult{41.26 / 0.97 / 5.22}&\bestResult{36.48 / 0.93 / 3.43} \\
		\hline
	\end{tabular}
	}
	\caption{Quantitative evaluation of the state-of-the-art SISR and RefSR algorithms, in terms of PSNR/SSIM/IFC for scale factors $\times 4$ and $\times 8$ respectively.}
	\label{Table:main_experiment}
\end{table*}

\begin{figure}[!ht]
	\centering
	\subfloat[][results on Flower$\times4$.]{\includegraphics[width=.45\textwidth]{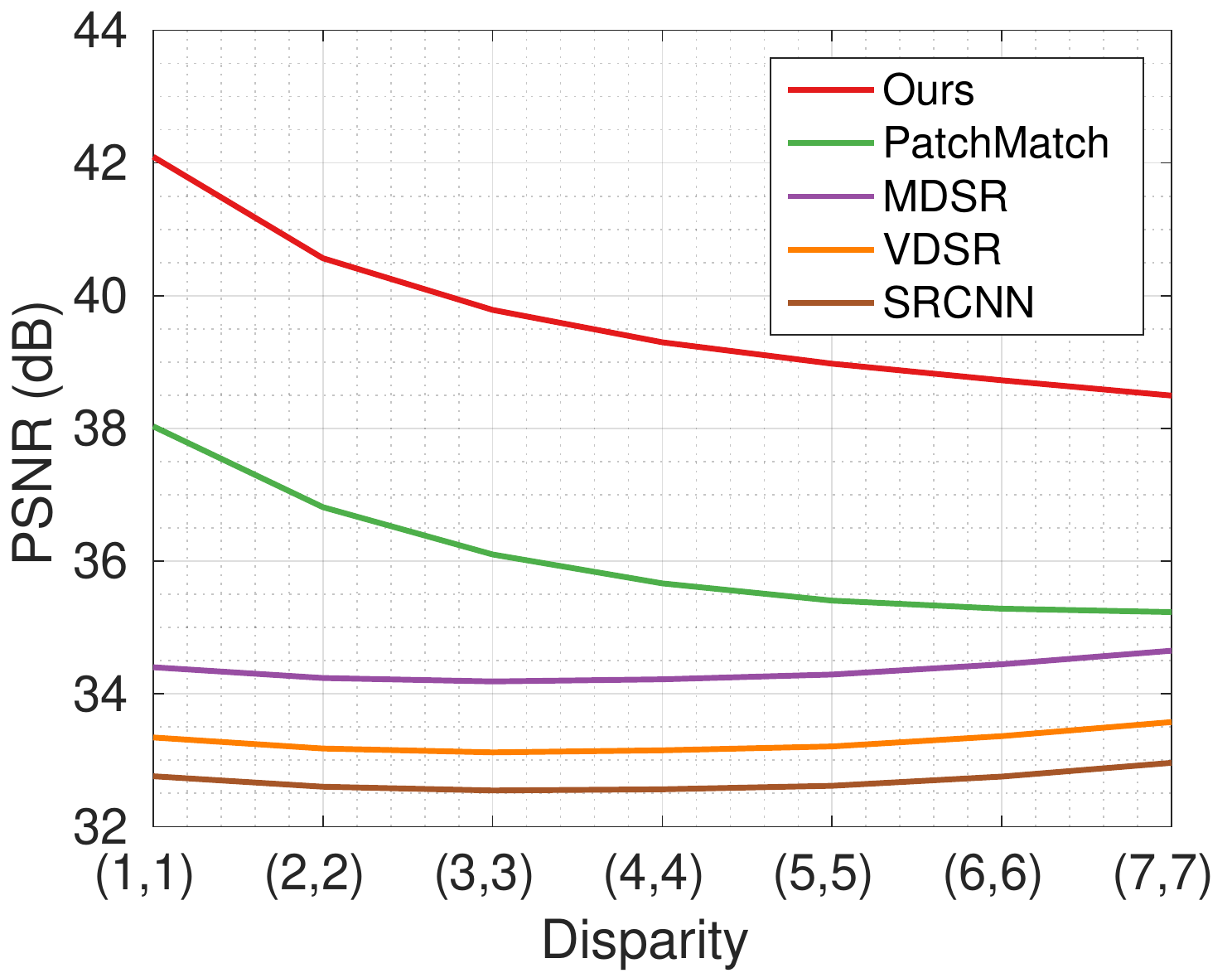}}\quad
	\subfloat[][results on LFVideo$\times4$.]{\includegraphics[width=.45\textwidth]{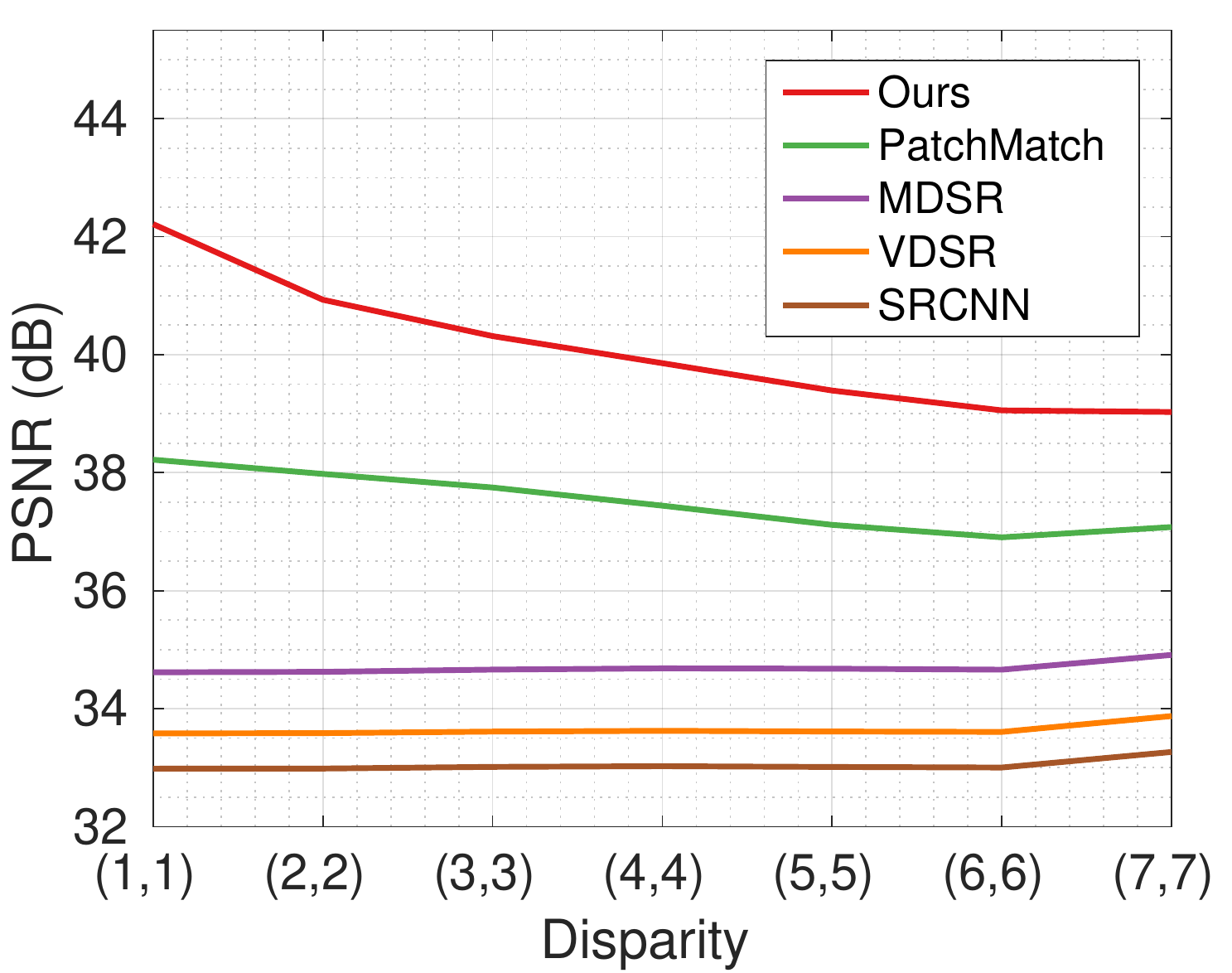}}\\
	\subfloat[][results on Flower$\times8$.]{\includegraphics[width=.45\textwidth]{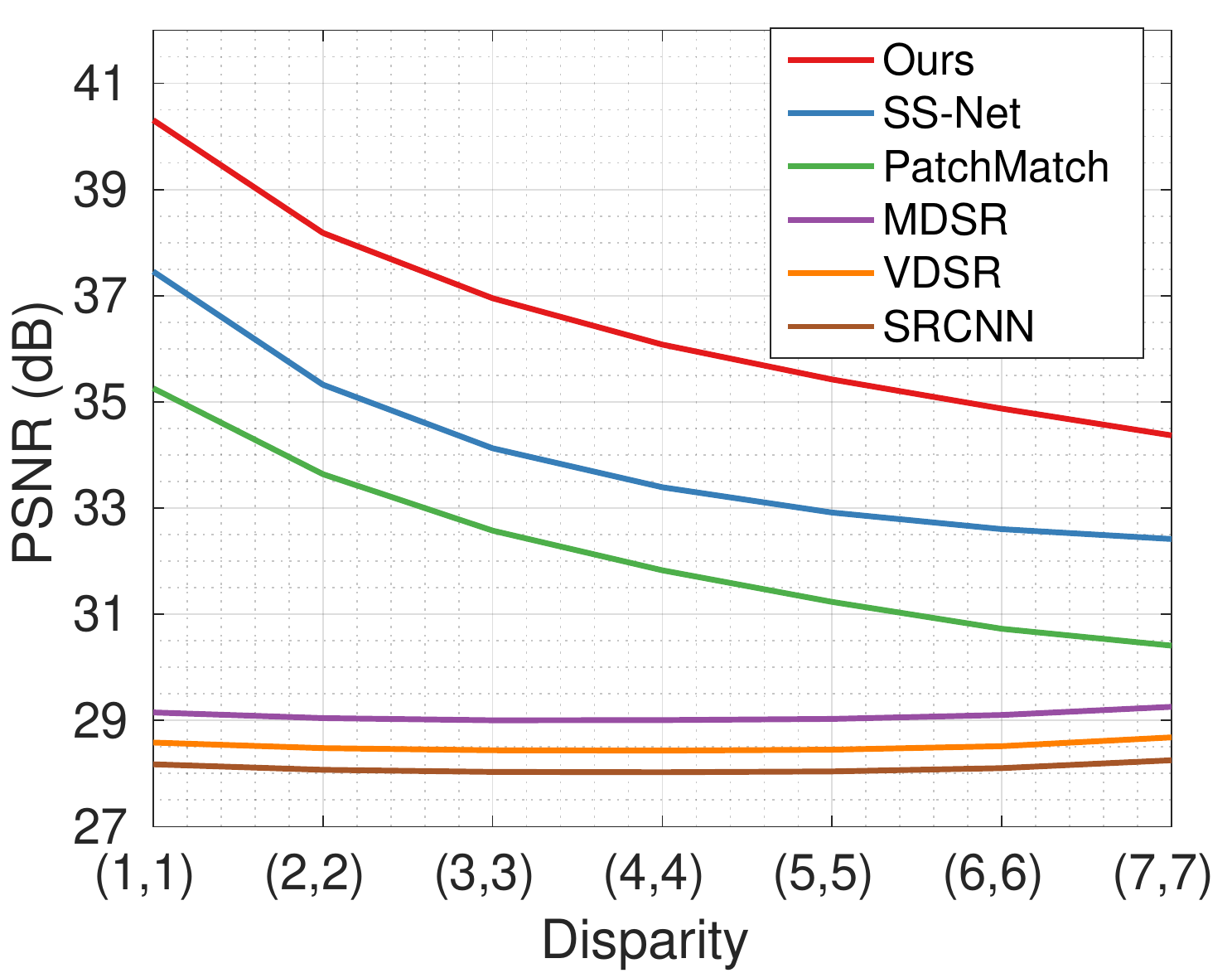}}\quad
	\subfloat[][results on LFVideo$\times8$.]{\includegraphics[width=.45\textwidth]{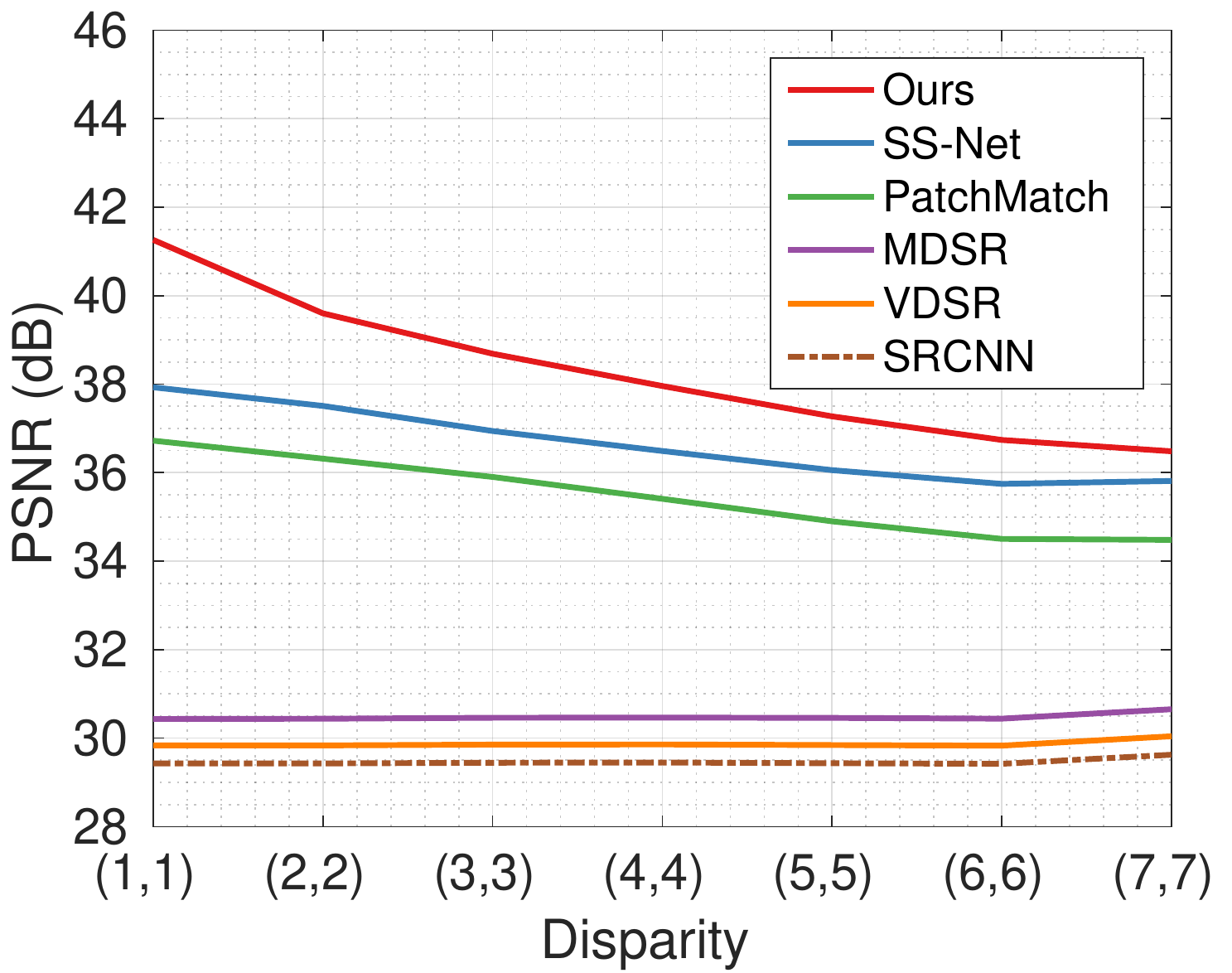}}
	\caption{The PSNR measurement under different parallax settings: the reference images are select at $(0,0)$ LF grid, while the LR image are selected at $(i,i)$ LR grid ($(i,i), 0 < i \le 8$).}
	\label{fig:PSNR_curve}
\end{figure}

For better comparison, we also visualize the PSNR performance under different parallax setting in Fig. \ref{fig:PSNR_curve}. As expected, the RefSR approaches such as CrossNet, PatchMatch, SS-Net outperform SISR approaches owe to the high-frequency details provided by reference images. However, RefSR results deteriorate as the parallax enlarges, due to the fact that the correspondence searching is more difficult for large parallax. In contrast, the performance of SISR approaches appears as `U-shape' for different views, i.e., at the corners of LF image for disparity being (1, 1) and (7, 7), the SISR performs slightly better. This is probably due to the occurrence of easily super-resolved invalid region becomes larger at corners. Finally, it can be seen that the proposed CrossNet consistently outperforms the resting approaches under different disparities, datasets and scales.

Fig. \ref{Table:visual-comparison1} presents the visual comparisons of CrossNet with SISR approaches including SRCNN, VDSR, MDSR and RefSR approaches including PatchMatch and SS-Net under the challenging $\times 8$ scale setting. Benefiting from the reference image, RefSR approaches show competitive results compared to the SISR methods, where the high frequency details are explicitly retained. Among them, the proposed CrossNet can further provide finer details, resembling the details in ground truth image. \zht{More visual comparison are shown in the supplementary material and supplementary video \footnote{\url{https://youtu.be/7htEaaNkxG8}}.}
\newcommand{\figw}{0.26}
\newcommand{\drawfigure}[1]{
\begin{minipage}[c]{1.0\textwidth}
	\begin{minipage}[c]{.35\textwidth}
		\vspace*{9px}
		\subfloat{
			\stackunder[5pt]{\includegraphics[width=1.0\linewidth]{#1/GT_all.png}}{Ground-truth HR}
		}		
	\end{minipage} 
	\begin{minipage}[b]{.45\textwidth}
		\begin{tabular}[c]{c c c c}
			\subfloat{
				\stackunder[5pt]{\includegraphics[width=\figw\linewidth]{#1/LR.png}}{LR}}		&
			\subfloat{
				\stackunder[5pt]{\includegraphics[width=\figw\linewidth]{#1/SRCNN.png}}{SRCNN\cite{SRCNN}}}		&
			\subfloat{
				\stackunder[5pt]{\includegraphics[width=\figw\linewidth]{#1/VDSR.png}}{VDSR\cite{VDSR}}}		&
			\subfloat{
				\stackunder[5pt]{\includegraphics[width=\figw\linewidth]{#1/MDSR.png}}{MDSR\cite{MDSR}}
			}	\\
			\subfloat{
				\stackunder[5pt]{\includegraphics[width=\figw\linewidth]{#1/PM.png}}{PatchMatch\cite{PatchMatch}}}		&
			\subfloat{
				\stackunder[5pt]{\includegraphics[width=\figw\linewidth]{#1/SSNet.png}}{SS-Net\cite{SS-Net}}}		&
			\subfloat{
				\stackunder[5pt]{\includegraphics[width=\figw\linewidth]{#1/EWDNet.png}}{Ours}}		&
			\subfloat{
				\stackunder[5pt]{\includegraphics[width=\figw\linewidth]{#1/GT.png}}{GT}}
			
		\end{tabular}
	\end{minipage}
\end{minipage}
}

\begin{figure}[!htb]
	\setlength{\belowcaptionskip}{-10pt}
	\centering
	\begin{tabular}[t]{c}
		\begin{tabular}[c]{c}
			\drawfigure{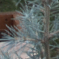} \\
			\drawfigure{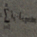} \\
			\drawfigure{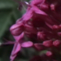} \\
		\end{tabular}
	\end{tabular}
	\caption{Visual comparison for $\times 8$ RefSR on LFVideo$(1,1)$, LFVideo$(3,3)$, Flower$(1,1)$. In the experiment, our approach is compared against SRCNN\cite{SRCNN}, VDSR\cite{VDSR}, MDSR\cite{MDSR}, PatchMatch\cite{PatchMatch}, and SS-Net\cite{SS-Net}.  }
	\label{Table:visual-comparison1}
\end{figure}

{\bf Generalization:}\label{Sec:ExpGenera} to further estimate the cross-dataset generalization capacity of our model, we report the results on Stanford light field dataset (Lego Gantry) \cite{Stanford-light-field} and the Scene Light Field dataset \cite{Scene_LF}, where the former one contains light field images captured by a Canon Digital Rebel XTi that set on a movable Mindstorms motors on the Lego gantry, and images from the latter one are also captured on a motorized stage with a standard DSLR camera. Under such equipment settings, the captured light-field images of the two datasets have much large parallax comparing to the ones captured by Lytro ILLUM cameras. The parallax discrepancy between datasets yields difficulty to our trained model, as our model is not particularly trained with large parallax.

To handle these two datasets, we employ a parallax augmentation procedure during training, which randomly offsets the reference input by $[-15,15]$ pixels both horizontally and vertically. We take the pre-trained model parameters using LFVideo dataset (in Section \ref{subsection:CrossNet}) as the initialization, and re-train the CrossNet on the Flower dataset for 200K iterations in order to achieve better generalization. We use 7e-5 as the initial learning rate, and decay the learning rate using factors 0.5, 0.2, 0.1 at 50K, 100K, 150K iterations.

Table \ref{Table:Stanford-dataset} and Table \ref{Table:SceneLF-dataset} compare in PSNR measurement our re-trained model with PatchMatch \cite{PatchMatch}, SS-Net \cite{SS-Net} for $\times8$ RefSR on the Stanford light-field dataset and the Scene Light Field dataset respectively. It can be seen that  our approach outperforms the resting approaches with different parallax settings on the Stanford dataset. On average, our approach outperforms the competitive SS-Net by 1.79-2.50dB on the Stanford light-field dataset and 2.84dB on the Stanford light-field dataset.

\begin{table}[ht!]
	\centering
	\small
	\resizebox{\textwidth}{!}{%
		\centering
		\begin{tabular}{|l|c|c|c|c|c|c|}
			\hline
			Image, parallax=(1/3/5,0)  			& MDSR \cite{MDSR} & PatchMatch \cite{PatchMatch} & SS-Net \cite{SS-Net} &Ours \\
			\hline
			$Amethyst$ 				& 29.64 / 29.66 / 29.74 & 36.44 / 34.71 / 33.20 &36.91 / 34.97 / 33.35 &\bestResult{39.52} / \bestResult{36.92} / \bestResult{35.13} \\
			$Bracelet$ 				& 24.66 / 24.68 / 24.64 & 35.66 / 33.71 / 25.47 &36.33 / 34.19 / \bestResult{32.53} &\bestResult{38.19} / \bestResult{34.27} / 27.20 \\
			$Chess$  				& 30.39 / 30.42 / 30.39 & 38.68 / 36.68 / 34.99 &39.85 / 38.64 / 37.12 &\bestResult{41.85} / \bestResult{40.68} / \bestResult{39.34} \\
			$Flowers$ 				& 30.01 / 30.00 / 29.98 & 33.74 / 33.24 / 32.58 &37.46 / 35.44 / 34.09 &\bestResult{39.50} / \bestResult{36.53} / \bestResult{34.56} \\
			$Jelly Beans$  			& 41.09 / 41.00 / 41.15 & 39.48 / 38.68 / 37.19 &37.98 / 36.60 / 35.14 &\bestResult{43.81} / \bestResult{42.29} / \bestResult{40.11} \\
			$Lego Bulldozer$  		& 29.58 / 29.56 / 29.58 & 35.60 / 31.39 / 28.87 &35.99 / 33.26 / 31.86 &\bestResult{38.79} / \bestResult{35.00} / \bestResult{32.61} \\
			$Lego Gantry$ 			& 26.58 / 26.52 / 26.58 & 31.73 / 29.86 / 27.15 &32.68 / \bestResult{30.97} / \bestResult{30.06} &\bestResult{33.42} / 30.83 / 29.96 \\
			$Lego Knights$  		& 29.49 / 29.48 / 29.46 & 33.57 / 30.73 / 27.57 &33.48 / 31.45 / 30.20 &\bestResult{37.60} / \bestResult{34.40} / \bestResult{32.11} \\
			$Lego Truck$ 			& 30.82 / 30.80 / 30.67 & 34.96 / 34.22 / 33.30 &37.80 / 36.44 / 34.87 &\bestResult{39.87} / \bestResult{38.51} / \bestResult{37.17} \\
			$Tarot Cards Large$  	& 22.91 / 22.89 / 22.86 & 27.98 / 20.90 / 20.40 &29.69 / \bestResult{26.71} / \bestResult{24.63} &\bestResult{31.27} / 23.69 / 22.16 \\
			$Tarot Cards Small$  	& 23.98 / 23.98 / 23.97 & 30.08 / 29.30 / 27.60 &32.92 / \bestResult{31.44} / \bestResult{30.70} &\bestResult{35.48} / 31.42 / 28.22 \\
			$Stanford Bunny$ 		& 36.82 / 36.90 / 36.96 & 37.39 / 37.15 / 36.75 &40.36 / 39.77 / 39.09 &\bestResult{41.99} / \bestResult{41.48} / \bestResult{40.88} \\
			\hline                                                                                                           
			Average 				& 29.66 / 29.66 / 29.67 & 34.61 / 32.55 / 30.42 &35.96 / 34.16 / 32.81&\bestResult{38.44} / \bestResult{35.50} / \bestResult{33.29} \\
			\hline											
		\end{tabular}
	}
	\caption{$\times 8$ super-resolution experiment on the Stanford light field dataset \cite{Stanford-light-field}.}
	\label{Table:Stanford-dataset}
	\centering
	\small
		\centering
		\begin{tabular}{|l|c|c|c|c|c|}
			\hline
			Image, parallax=(1,0)& MDSR \cite{MDSR} & PatchMatch \cite{PatchMatch} & SS-Net \cite{SS-Net} &Ours \\
			\hline
			$Bikes$ 				& 28.38  & 36.70  & 36.36  & \bestResult{37.91} \\
			$Church$ 				& 36.04  & 41.89  & 40.10  & \bestResult{43.68} \\
			$Couch$  				& 32.52  & 33.93  & 35.86  & \bestResult{39.83} \\
			$Mansion$ 				& 28.03  & 32.83  & 33.39  & \bestResult{36.38} \\
			$Statue$  				& 29.72  & 35.96  & 35.21  & \bestResult{37.30} \\
			\hline
			Average 				& 30.94  & 36.26  & 36.18  & \bestResult{39.02} \\
			\hline										
		\end{tabular}
	\caption{$\times 8$ super-resolution experiment on the Scene light field dataset \cite{Scene_LF}.}
	\label{Table:SceneLF-dataset}
\end{table}

{\bf Efficiency}: It is worth mentioning that the proposed CrossNet generates an $320\times512$ image for $\times8$ RefSR within 1 second, i.e., 0.75 second to perform SISR preprocessing using the MDSR \cite{MDSR} model, and 0.12 seconds to synthesize the final output. In contrast, the PatchMatch \cite{PatchMatch} takes 86.3 seconds to run on Matlab2016 using GPU parallelization while the SS-Net \cite{SS-Net} takes on average 105.6 seconds running on GPU. The above running times are profiled using a machine with 8 Intel Xeon CPU (3.4 GHz) and a GeForce GTX 1080 GPU, while the model inferences of our CrossNet and SS-Net \cite{SS-Net} are implemented on Python with Theano deep learning package \cite{theano}.
\vspace{-0.3cm}

\subsection{Discussions}
\label{Sec:ExpDis}
\begin{figure}[htbp]
	\begin{minipage}[b]{0.45\linewidth}
		\includegraphics[width=6cm,height=4cm]{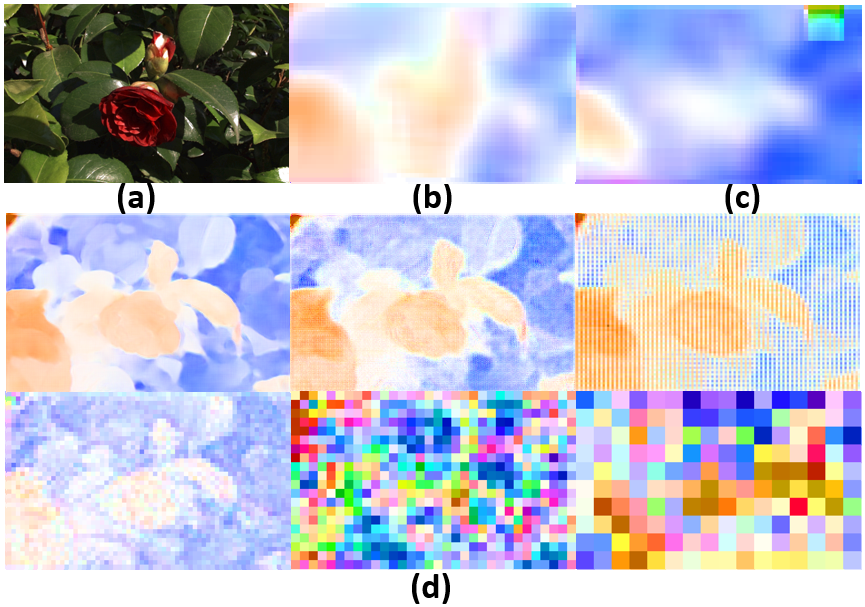}
	\end{minipage} ~~~~~~
	\begin{minipage}[b]{0.45\linewidth}
		\includegraphics[width=6cm,height=4cm]{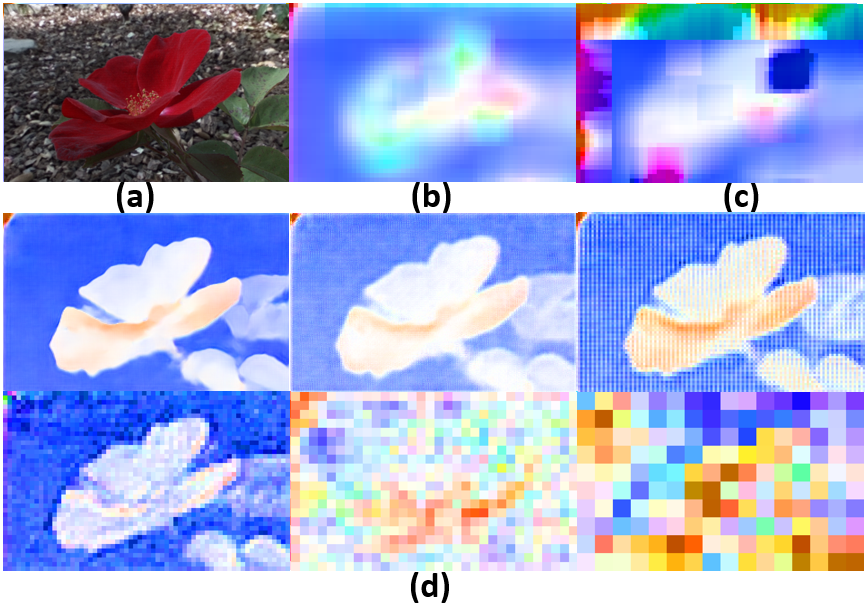}
	\end{minipage}
	\vspace{-10pt} 
	\caption{Flow visualization and comparison for sample \#1, \#99 in the Flower $\times 8$ testing set. (a) the HR image, (b)(c)(d) flow visualization of PatchMatch~\cite{PatchMatch}, SS-Net \cite{SS-Net}, and our approach respectively. In (d), the flow is visualized at scales $\times 1,\times 2,\times 4$ (row 1), and $\times 8,\times 32,\times 64$ (row 2).} 
	\label{fig:flow_visualize}
\end{figure}

One may concern that our loss is designed for image synthesis, and does not explicitly define terms for flow estimation. However, since the correctly aligned features are extremely informative for decoder to reconstruct high-frequency details, our model actually learns to predict optical flow by aligning features maps in an unsupervised fashion. To validate the effectiveness of the learned flow by aligning feature, we visualize the intermediate flow field at all scales in Fig. \ref{fig:flow_visualize}(d), where flow predictions at scales $0,1,2,3 (\times 1,\times 2,\times 4,\times 8)$ are reasonably coherent, yet noisy flow predictions are observed at scales $4,5 (\times 16,\times 32)$, because the flow at scale $4,5$ are not used for the feature-domain warping.

In addition to the multi-scale feature warping module proposed in this paper, we investigate a single-scale image warping counterpart which performs reference image warping \textbf{before} the following image encoder for feature extraction. This counterpart is inspired by the common practice in \cite{LFVideo,Learning-based-view-synthesis} that performs image warping before synthesis. More concretely, our single-scale image warping counterpart performs image warping using the flow from scale $0$: $\hat{I}_{REF} = warp(I_{REF}, V^{(0)}).$ After that, reference image encoder with the same structure is used to extract features from the warped reference image. Without changing the structure of encoder and decoder, such image warping counterpart CrossNet-iw has the same model size as CrossNet.

\begin{figure}[!ht]
	\centering
	\subfloat[][Training convergence curve on Flower$\times8$.]{\includegraphics[width=.45\textwidth]{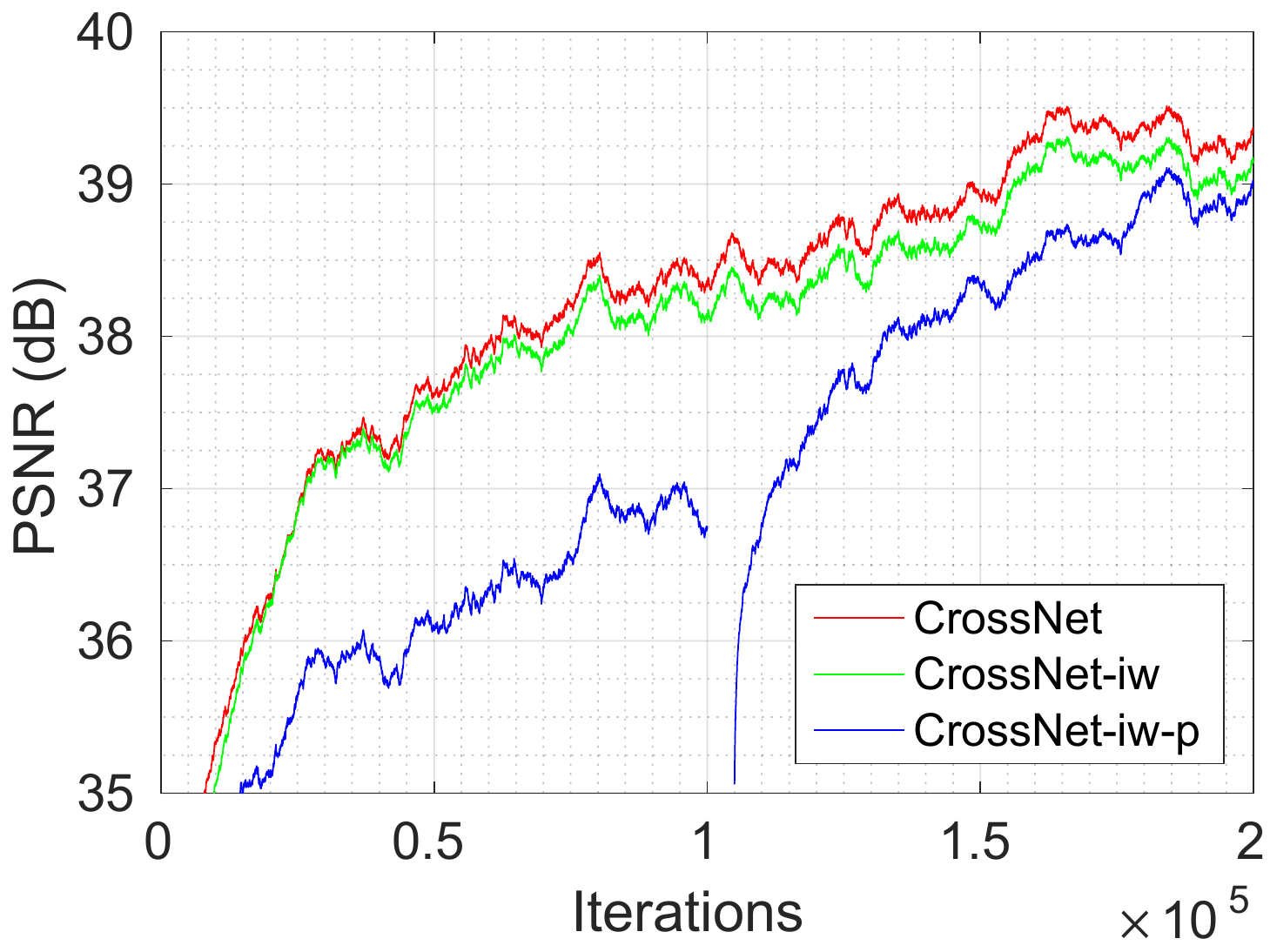}}\quad
	\subfloat[][Training convergence curve on LFVideo$\times8$.]{\includegraphics[width=.45\textwidth]{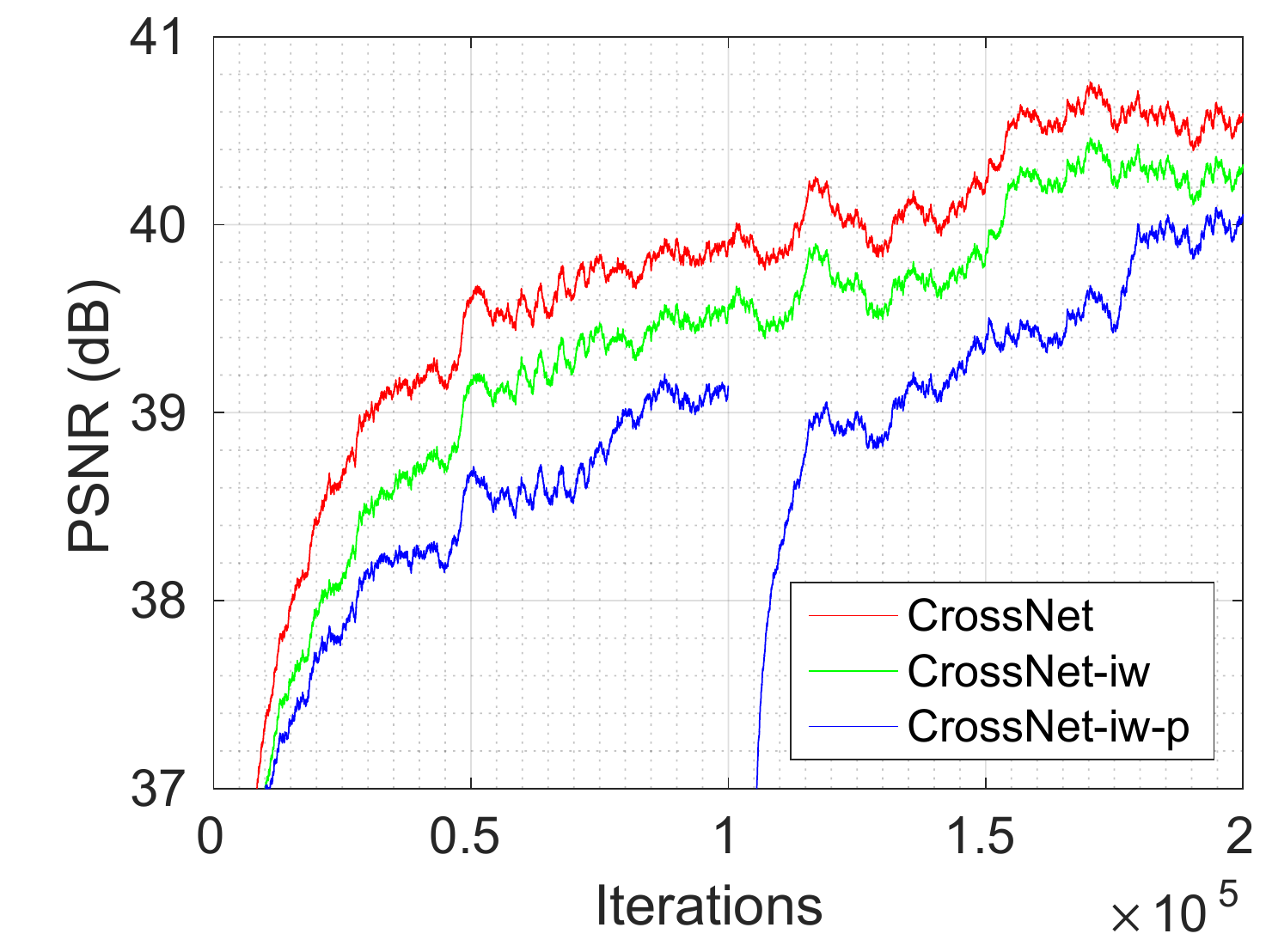}}\\
	\caption{\zht{The convergence analysis on our feature domain warping scheme (CrossNet) versus image warping schemes. Our model (CrossNet, red) converges faster than the image-domain warping counterpart (CrossNet-iW) with or without pre-training.}}
	\label{fig:converge_curve}
\end{figure}

We train both CrossNet-iw and CrossNet according to the same procedure in Section \ref{subsection:CrossNet}. We also adopt a pretraining strategy to train CrossNet-iw. We pretrain the flow estimator of WS-SRNet with image warping task for 100K iterations, and then apply the joint training for another 100K iterations, resulting the CrossNet-iw-p model. Fig. \ref{fig:converge_curve} shows the PSNR convergence curves on training set for $\times8$RefSR on the Flower and LFVideo dataset. It can be noticed that our CrossNet converges faster than the CrossNet-iw counterparts. At the end of the training, CrossNet outperforms CrossNet-iw 0.20dB and 0.27dB on training set. Table \ref{Table:ablation_feature_warp} shows the RefSR precision on the test sets with three representative point views. CrossNet outperforms CrossNet-iw, especially on small parallax setting. It is reasonable because the training uses random sampled pairs from the LF grid, which are mostly took up by small parallax training pairs.

\begin{table*}[htbp]	
	\centering
	\begin{tabular}{ |l|c|c|c|c|c| }
		\hline
		Model 								& parameter size&  Flower$\times 8$ (1,1) & Flower$\times 8$ (3,3) & Flower$\times 8$ (7,7) \\
											&  			 	&  PSNR/SSIM/IFC 			& PSNR/SSIM/IFC			  & PSNR/SSIM/IFC\\
		\hline
		\bestResult{CrossNet}				& 41M 	 		& \bestResult{40.31 / 0.98 / 5.74}		&\bestResult{36.95 / 0.96 / 4.61}	&34.37 / 0.93 / 3.45\\
		CrossNet-iw							& 41M 	 		& 40.14 / 0.98 / 5.75					&36.94 / 0.96 / 4.59				&\bestResult{34.49 / 0.93 / 3.52}\\
		CrossNet-iw-p						& 41M 	 		& 40.01 / 0.98 / 5.75 					&36.85 / 0.96 / 4.58				&34.35 / 0.93 / 3.49\\
		\hline
		  									&  				&  LFVideo$\times 8$ (1,1) & LFVideo$\times 8$ (3,3) & LFVideo$\times 8$ (7,7) \\
											&  			 	&  PSNR/SSIM/IFC 			& PSNR/SSIM/IFC			  & PSNR/SSIM/IFC\\
		\hline
		\bestResult{CrossNet}				& 41M 	 		& \bestResult{41.26 / 0.97 / 5.22}		&\bestResult{38.69 / 0.96 / 4.32}	&\bestResult{36.48 / 0.93 / 3.43}\\
		CrossNet-iw							& 41M 	 		& 41.12 / 0.97 / 5.20					&38.61 / 0.96 / 4.32 				&36.32 / 0.93 / 3.43\\
		CrossNet-iw-p						& 41M 	 		& 40.96 / 0.97 / 5.16 					&38.47 / 0.96 / 4.29				&36.11 / 0.93 / 3.43\\
		\hline
	\end{tabular}
	\caption{Ablation study to evaluate the effectiveness of multi-scale feature warping.}
	\label{Table:ablation_feature_warp}
\end{table*}

\vspace{-0.7cm}
As our method relies on the cross-scale flow estimators, it is also important to study the flow predicting capacities of different flow estimator. For such purpose, we train the FlowNetS and our modified model (FlowNetS+) on the Flower and the LFVideo dataset for warping the reference images to the ground truth images given the reference and LR image as input. As shown In Table \ref{Table:flow-estimator}, while the FlowNetS+ contains 2\% more parameters in comparison to FlowNetS, the additional upscaling layers of FlowNetS+ reasonably improves the warping precision in both the Flower dataset \cite{Flower} and the LFVideo dataset \cite{LFVideo}, as they help to generate finer flow field. In addition, we also observe that by plain warping, the FlowNetS+ achieves notably better compatible performance compared to SS-Net \cite{SS-Net}, as depicted by the SS-Net ($\times8$) row in Table \ref{Table:main_experiment}.
\vspace{-0.6cm}
\begin{table}[htbp]
	\resizebox{\textwidth}{!}{%
		\centering
		\begin{tabular}{ |l|c|c|c|c|c| }
			\hline
			Model 							& \# of parameters & Flower$\times8$(1,1)  & Flower$\times8$(7,7)  & LFVideo$\times8$(1,1) & LFVideo$\times8$(7,7) \\
											&	    		& PSNR/SSIM/IFC	    	& PSNR/SSIM/IFC			& PSNR/SSIM/IFC			& PSNR/SSIM/IFC 		\\
			\hline
			FlownetS						& 31.9 million 		&37.78 / 0.97 / 5.41	& 31.23 / 0.90 / 3.02	&\bestResult{39.39 / 0.97 / 4.97}	&34.94 / 0.92 / 3.30 \\
			{\bf FlownetS+}						& 32.6 million 		&\bestResult{38.04 / 0.97 / 5.46} 	& \bestResult{31.66 / 0.90 / 3.11}	&{39.37 / 0.97 / 4.88}	&\bestResult{35.85 / 0.93 / 3.54} \\
			\hline
		\end{tabular}
	}
	\caption{Quantitative evaluation and the parameter sizes comparison, using different flow estimators to warp the reference image. The LR images are located at angular position $(3,3)$.}
	\label{Table:flow-estimator}
\end{table}

\vspace{-1.4cm}

%% file: section/conclude.tex
\section{Conclusion}
\vspace{-0.3cm}
Aiming for the challenge large-scale ($8\times$) super-resolution problem, we propose an end-to-end reference-based super resolution network named as CrossNet, where the input is a low-resolution (LR) image and a high-resolution (HR) reference image that shares similar view-point, the output is the super-resolved (4x or 8x) result of LR image. The pipeline of CrossNet is full-convolutional, containing encoder, cross-scale warping, and decoder respectively. Extensive experiment on several large-scale datasets demonstrate the superior performance of CrossNet (around 2dB-4dB) compared to previous methods. More importantly, CrossNet achieves a speedup of more than 100 times compared to existing RefSR approaches, allowing the model to be applicable for real-time applications.

\newpage